\documentclass[11pt,a4paper]{article}
\pdfoutput=1
\usepackage[hyperref]{naaclhlt2019}
\usepackage{times}
\usepackage{latexsym}
\usepackage{amsmath}
\usepackage{graphicx}
\usepackage{xcolor}
\usepackage{todonotes}
\usepackage{tabularx}
\usepackage{multirow}

\DeclareMathOperator*{\argmax}{argmax}

\usepackage{color}

\newcommand{\malopa}{{M{\sc a}LOP{\sc a}}}

\usepackage{url}

\aclfinalcopy 
\def\aclpaperid{1894} 

\title{A Grounded Unsupervised Universal Part-of-Speech Tagger for Low-Resource Languages}

\author{Ronald Cardenas$^\clubsuit$ \enspace Ying Lin$^\spadesuit$ \enspace Heng Ji$^\spadesuit$ \enspace Jonathan May$^\heartsuit$\\
  $^\clubsuit$ Institute of Formal and Applied Linguistics, Charles University in Prague \\
  $^\spadesuit$ Computer Science Department, Rensselaer Polytechnic Institute \\
  $^\heartsuit$ Information Sciences Institute, University of Southern California \\
  {\tt ronald.cardenas@matfyz.cz liny9@rpi.edu}\\ 
  {\tt jih@rpi.edu jonmay@isi.edu} 
  }

\date{}

\begin{document}
\maketitle

\begin{abstract}
Unsupervised part of speech (POS) tagging is often framed as a clustering problem, but practical taggers need to \textit{ground} their clusters as well. Grounding generally requires reference labeled data, a luxury a low-resource language might not have. In this work, we describe an approach for low-resource unsupervised POS tagging that yields fully grounded output and requires no labeled training data. We find the classic method of \newcite{brown1992class} clusters well in our use case and employ a decipherment-based approach to grounding. This approach presumes a sequence of cluster IDs is a `ciphertext' and seeks a POS tag-to-cluster ID mapping that will reveal the POS sequence.  
We show intrinsically that, despite the difficulty of the task, we obtain reasonable performance across a variety of languages. We also show extrinsically that incorporating our POS tagger into a name tagger leads to state-of-the-art tagging performance in Sinhalese and Kinyarwanda, two languages with nearly no labeled POS data available.
We further demonstrate our tagger's utility by incorporating it into a true `zero-resource' variant of the {\malopa} \cite{malopa} dependency parser model that removes the current reliance on multilingual resources and gold POS tags for new languages. Experiments show that including our tagger makes up much of the accuracy lost when gold POS tags are unavailable.

\end{abstract}

\section{Introduction}
While cellular, satellite, and hardware advances have ensured that sophisticated NLP technology can reach all corners of the earth, the language barrier upon reaching remote locales still remains. As an example, when international aid organizations respond to new disasters, they are often unable to deploy technology to understand local reports detailing specific events \cite{munromanning2012,crisis-mt-developing-a-cookbook-for-mt-in-crisis-situations}. An inability to communicate with partner governments or civilian populations in a timely manner leads to preventable casualties.

\begin{figure}[!t]
\centering
\includegraphics[width=0.85\linewidth]{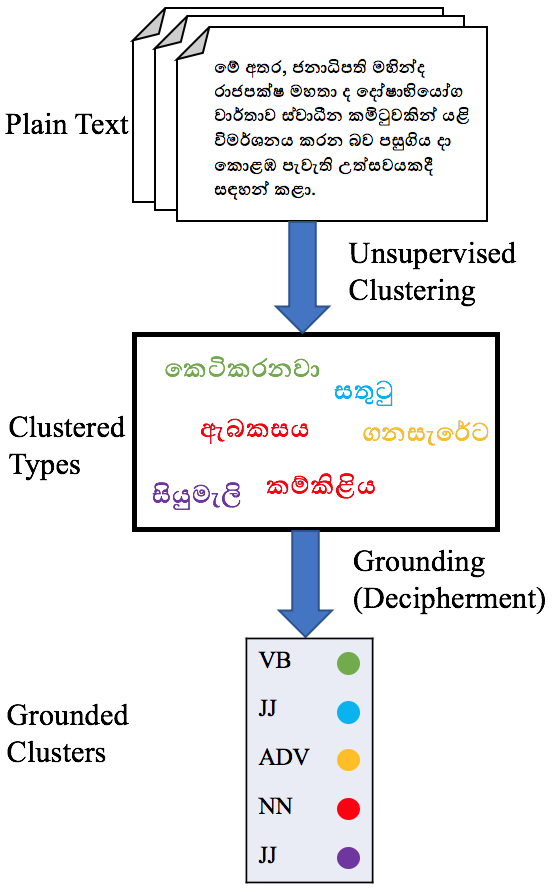}
\caption{Overview of our approach to grounded POS tagging. We use an unsupervised clustering method (Section~\ref{sec:clust}) then reduce and ground the clusters using a decipherment approach informed by POS tag sequence data from many languages (Section~\ref{sec:grounding}).}
\label{figure:overview}
\end{figure}

The lack of adequate labeled training data has been the major obstacle to expanding NLP's outreach more multilingually. Developments in unsupervised techniques that require only monolingual corpora \cite{lample2017unsupervised,artetxe2017unsupervised} and the ability to leverage labeled resources in other languages  have been proposed to address this issue \cite{das2011unsupervised,duong-EtAl:2014:EMNLP2014,malopa}. Unfortunately, these methods either do not work in practice on true low-resource cases or unrealistically assume the availability of some amount of supervision.

Consider syntactic parsing as a prime example. 
Past editions of the CoNLL Shared Task on Multi-lingual Parsing \cite{zeman2017conll,zeman2018conll} featured a category of target languages for which either little or no training data was provided. However, even in the `no-resource' scenario that most closely matches our use case, gold part-of-speech (POS) tags for test data were provided for the participants to use. Prior to these shared tasks, \citet{malopa} proposed a variant of their main model, \malopa, that was meant to produce reasonable parses for languages under ``zero-resource'' conditions.
In order to function, however, the model requires users to provide gold POS tags and word mappings from these languages into a common semantic space, using approaches that require parallel data \cite{guo2015cross}.

Indeed, the compulsion to use POS tag-labeled data in zero-resource circumstances extends to the vast, varied lines of research in unsupervised POS tagging itself! Every approach explored so far ultimately requires POS-annotated resources for the language being studied in order to produce a final, grounded output. Even the most conservative strategies \cite{goldwater2007fully,berg2010painless,stratos2016unsupervised} that do not require any supervised signal during training still ultimately produce only ungrounded clusters, and require a reference annotated corpus to map the inferred clusters or states to actual POS tags.

Making matters worse, evaluation is generally offered in terms of the `many-to-one' or `one-to-one' analyses \citet{johnson2007doesn}. These metrics use a reference corpus to determine the \textit{optimal} mapping of clusters to tags. While this evaluation approach is intuitively sensible for measuring cluster purity, to actually \textit{use} such an output, an entire annotated training corpus is required.\footnote{Additionally, \newcite{headdeniii-mcclosky-charniak:2008:PAPERS} demonstrated that these metrics are not indicative of downstream performance.} It is not enough to simply rely on ungrounded clusters in real-world systems; grounded labels offer a sort of universal API between other resources such as rule-based modules that operate on certain word types or between resources built from other annotated high-resource language data. 

Since POS tag and parallel data resources for new languages are often unavailable or unreliable, we make the following contributions to ensure the surprise of a new language does not immobilize us:
\begin{itemize} \itemsep -1pt
    \item We introduce a decipherment-based approach to POS grounding, which yields fully grounded output and does not require any annotated data or parallel corpora in the language to be analyzed. The approach uses pre-existing human-labeled POS tag sequences from high-resource \textit{parent languages} (PL) but no labeled data or sequences for the target, or \textit{child language} (CL). An overview of the approach is shown in Figure~\ref{figure:overview}.
    \item We demonstrate our approach by evaluating over a variety of languages spanning 4 families and 8 genera (Germanic, Romance, Slavic, Japanese, Semitic, Iranian, Indic, and Bantoid), and show across-the-board reasonable intrinsic performance, given the difficulty of the task and the stringency (straightforward accuracy) in comparison to other unsupervised evaluation strategies.
    \item We test the utility of our grounded tags in a name tagging task, obtaining state-of-the-art performance for Sinhalese and Kiryarwanda, two languages with nearly no labeled POS or named entity resources.
    \item We further pare down the annotated resources required in an existing `zero-resource' dependency parser model and show that our unsupervised and grounded tags are helpful at closing the gap between a nihilistic tag-free setting and an unrealistic gold tag setting.
    \item We release our code so that others may create zero-resource syntactic analysis and information extraction systems at the onset of the next new emergency.\footnote{\url{https://github.com/isi-nlp/universal-cipher-pos-tagging.git}}
\end{itemize}

\section{POS Grounding as Decipherment}
\label{section-prob}

We consider the task of POS induction as a two-step pipeline: from word sequence $w$ to POS tag sequence $p$ via cluster sequence $c$. Formally, our conditional probability model is

\begin{align*}
    \argmax_p & P_\theta(p|w) \\ & =  \argmax_p \sum_{c \in C^{|w|}} P_\theta(p,c|w) \\
     & = \argmax_p \sum_{c \in C^{|w|}} P_\theta(p|c,w)P_\theta(c|w)
\end{align*}

\noindent where $C$ is the cluster vocabulary and $\theta$ parameterizes our probability model. If we assume a deterministic pipelined clustering of words and a tag labeling model that does not depend on words, then for chosen $\hat{c}$, this becomes
\begin{align}
    \nonumber \argmax_p & \sum_{c \in C^{|w|}} P_\theta(p|c,w)P_\theta(c|w) \\
    \nonumber = & \argmax_p P_\theta(p|\hat{c}) \\
    = & \argmax_pP_\theta(\hat{c}|p)P_\theta(p) \label{eq-dec}
\end{align}

We call this model the \textit{cipher grounder}. As presented it requires an estimate for $P_\theta(p)$ for the CL, which requires POS training data. Under the zero-resource scenario, we instead approximate $P_{\theta}(p)$ by the tag distribution of a PL. Then, the {\it cipher table} $P_{\theta}(\hat{c}|p)$ can be trained using a noisy-channel, expectation-maximization (EM)-based approach as in \citet{ravi2011deciphering}.

\section{POS Tagger construction}
\label{section:exps}

We approach the search for optimal components in the two-step pipeline outlined in Section~\ref{section-prob} in a cascaded manner. First, an optimal word clustering is determined by means of the many-to-one evaluation method. This method is explained well by \newcite{johnson2007doesn}:

\begin{quote}
  `` ...deterministically map each hidden state to the POS tag it co-occurs most frequently with, and return the proportion of the resulting POS tags that are the same as the POS tags of the gold-standard corpus.''
\end{quote}

While unrealistic for POS tagger performance purposes, many-to-one is a good choice for determining cluster `purity' and provides a reasonable grounding upper bound. As the calculation of many-to-one does require labeled data, we constrain the use of these labels for development and will evaluate extrinsically using languages for which we do not have any training data; see Section~\ref{sec:extrinsic}. 

Secondly, we search for the best approach to ground the chosen clusters, given several possible PL options. 

After the optimal components and parameters are determined, we validate POS tag quality intrinsically via tag accuracy on reference data where it exists, and then extrinsically on two downstream tasks. We investigate a simulated no-resource scenarios in the task of dependency parsing, and a real low-resource scenario in name tagging.

\subsection{Datasets}

For intrinsic evaluation and optimization of the tagging pipeline, including  all preliminary experiments, we use annotated corpora from Universal Dependencies (UD) v2.2\footnote{\url{http://universaldependencies.org/}} for the following languages: English (en), German (de), French (fr), Italian (it), Spanish (es), Japanese (ja), Czech (cs), Russian (ru), Arabic (ar), and Farsi (fa). For Swahili (sw), we use the Helsinki Corpus of Swahili 2.0.\footnote{\url{http://urn.fi/urn:nbn:fi:lb-2016011301}}
Overall in these experiments we cover 11 languages and 4 language families.

In our dependency parsing experiments, we use the Universal Treebank v2.0 \cite{mcdonald2013universal} for en, de, fr, es, it, Portuguese (pt), and Swedish (sv). This set of treebanks is chosen instead of UD in order to obtain results comparable to those of previous work on simulated zero-resource parsing scenarios \cite{malopa,zhang2015hierarchical,rasooli2015density}.

In our name tagging experiments, we use monolingual texts for Sinhalese (si) and Kinyarwanda (rw) provided by DARPA's Low Resource Languages for Emergent Incidents (LORELEI) Program during the 2018 Low Resource Human Languages Technologies (LoReHLT) evaluation.

\subsection{Unsupervised Clustering}
\label{sec:clust}

In this step we compare two approaches to unsupervised ungrounded labeling. The first strategy is to cluster by word types and thus label each token with its cluster ID independently of its context.\footnote{We refer to ungrounded POS tag labels as `clusters' even though not all methods induce a clustering.} We consider Brown's hierarchical clustering algorithm, \cite{brown1992class}\footnote{\url{https://github.com/percyliang/brown-cluster}} {\sc brown}; Brown's exchange algorithm,\footnote{Optimized and implemented by \citet{marlin}. Available at\ \url{http://cistern.cis.lmu.de/marlin/}} \cite{MARTIN199819} {\sc marlin}; and k-means clustering of monolingual word embeddings of dimension size 100, trained using {\it fastText} \citep{joulin2016fasttext}, {\sc e-kmeans}.
The second labeling strategy is context-sensitive; it uses the Bayesian HMM tagger proposed by \citet{stratos2016unsupervised}, which we call {\sc a-hmm}. As noted previously, we evaluate unsupervised labeling extrinsically, via the many-to-one approach, and use the best performing labeling in the complete two-step grounded tagging pipeline. 

In preliminary experiments, we vary the number of clusters and hidden states ($|C|$) between 17 and 500. We initially sought to create one cluster per UD POS tag and then choose the proper 1:1 assignment of cluster to tag, following the approach of \newcite{stratos2016unsupervised}. However, cluster purity is low when only 17 clusters are allowed (i.e. each cluster has words with a variety of POS tags). Naturally, as the number of clusters is raised, the purity of each cluster improves. We ultimately fix the cluster limit at 500, which gives a good tradeoff between overall cluster quality for all the ungrounded tagging methods, and size small enough to allow EM-based decipherment to be tractable. 

Given this setting, we evaluate our four labeling strategies using the many-to-one approach, as presented in Table~\ref{table:m1-res}. Due to the larger number of clusters, the results presented here are higher than and not comparable to the original literature describing the methods.\footnote{As noted by \newcite{clark2003combining} and \newcite{johnson2007doesn}, in the limit, keeping each type (or, in the case of {\sc a-hmm, token} in its own cluster will result in the maximum possible many-to-one (polysemic types prevent perfect accuracy when type clustering).} We can, nevertheless, make relative judgements.  In all cases, clustering by type with Brown-based algorithms works better than using a sophisticated tagger such as {\sc a-hmm}.
Since {\sc brown} and {\sc marlin} obtain similar results, with no consistently dominant model, in all subsequent experiments we use the {\sc brown} labeler with 500 clusters.

\begin{table*}
\centering
\begin{tabular}{|l|cccccc|}
\hline
Seq. Tagger & en             & de             & fr            & ru             & fa             & sw             \\ \hline
{\sc brown}    & 81.37          & \textbf{81.28} & 84.81         & \textbf{79.78} & \textbf{86.94} & 87.35          \\
{\sc marlin}   & \textbf{81.53} & 81.25          & \textbf{85.4} & 79.14          & 86.64          & \textbf{88.81} \\
{\sc a-hmm}    & 77.12          & 74.85          & 81.48         & 73.88          & 80.25          & 76.69              \\
{\sc e-kmeans} & 63.01          & 65.14          & 68.68         & 70.80           & 76.94          & 65.08         \\ \hline
\end{tabular}
\caption{Comparison of labeling strategies using many-to-one mapping for target languages with available test data, using 500 clusters or number of states. Accuracy is shown in percentage points.}
\label{table:m1-res}
\end{table*}

\subsection{Grounding via Decipherment}
\label{sec:grounding}

We now seek an appropriate method for grounding the clusters generated in Section~\ref{sec:clust}. We experiment with en, fr, fa, and sw as CLs. For each CL $t$, we instantiate our model following Equation~\ref{eq-dec}, using the Carmel toolkit \citep{graehl1997carmel} and forming the cipher table as a one-state transducer. We train these models using EM for 500 iterations or until convergence, and we select the model with the lowest perplexity from among 70 random restarts.

Yet unspecified is the nature of the POS language model $P_\theta(p)$. We begin by training bigram models of POS tag sequences with additive smoothing using the SRILM toolkit \citep{stolcke2002srilm} for each PL $s \in \mathcal{S}$ = $\{$en, de, fr, it, es, ja, ar, cs, ru, sw$\}$. But which PL's POS tag data to use for each CL? We explore two initial criteria for choosing a single suitable PL $s$: confidence of the model during decoding (perplexity, {\sc PPL}), and typological similarity. For the first criterion, the PL whose cipher grounder $s$-$t$ yields the better performance is chosen. For the second criterion, the most similar language to CL $t$ is chosen according to the cosine similarity between typological features vectors. We employ 102 features obtained from WALS\footnote{\url{https://wals.info/}} related to word order and morphosyntactic alignment, further reduced to 50 dimensions using PCA. However, none these criteria correlates significantly to tagging accuracy, as we elaborate in Section~\ref{section:res-cipher}. We instead try a combined approach.

The likelihood of cluster ID replacement, $P_{\theta}(\hat{c}_i|p_j), \forall \hat{c}_i \in C, \forall p_j$ in the tagset, is replaced by 
\begin{align*}
P_{avg}(\hat{c}_i|p_j) \sim \frac{ \sum_{s \in \mathcal{S}, s \neq t} P_{\theta}(\hat{c}_i|p_j^s) }{|S|-1}
\end{align*}
where $P_{\theta}(\hat{c}_i|p_j^s)$ is the likelihood of POS tag $p_j$ being represented by cluster $\hat{c}_i$ after training with the language $s$ tag distribution.
Note that the CL is excluded from $\mathcal{S}$ for the combination.
The combined cipher grounder is then defined by
\begin{align}
\label{eq-dec-avg}
\argmax_{p} P_{all}(p) P_{avg}(\hat{c}|p)
\end{align}
where $P_{all}(p)$ is a language model trained over the concatenation of POS sequences of all parent languages in $\mathcal{S}$. We call this approach {\sc cipher-avg}.

\section{Downstream Tasks}

\subsection{Name Tagging}

\begin{figure}[!htb]
\centering
\includegraphics[width=\linewidth]{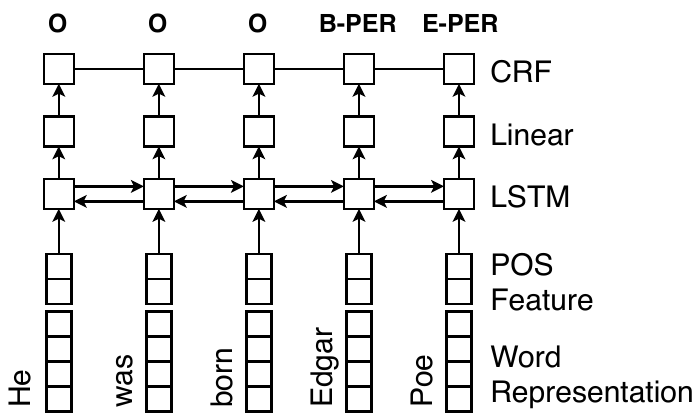}
\caption{Name tagging model evaluated in Section~\ref{sec:extrinsic}. This is an extension of the model of ~\newcite{chiu2016named} with POS tag features added.}
\label{figure:rnncrf}
\end{figure}

We experiment with the LSTM-CNN model proposed by \citet{chiu2016named}, one of the state-of-the-art name tagging models, as our baseline model. To incorporate POS features, we extend the token representation (word and character embeddings) with a one-hot vector representation of the POS tag. Figure \ref{figure:rnncrf} presents an outline of the architecture.

\subsection{Multilingual Dependency Parsing}

We base our experiments on the no-treebank setup of {\malopa}  \cite{malopa}, but change the underlying transition-based parser to the graph-based parser proposed by \citet{dozat2016biaffine} for implementation convenience.
Following this setup, for each CL except en, we train the parser on the concatenation of treebanks of the other 6 languages as PLs.

The original {\malopa} work enriches the input representation by concatenating pretrained multilingual word embeddings \cite{guo2016representation}, multilingual Brown cluster IDs, and POS tag information. However, these representations are obtained using parallel corpora and gold POS tags are required for parsing at test time.

In contrast, we are interested in the realistic scenario in which no resource is available in the child language but raw text.
It is important to note, however, that our objective is not to beat the state-of-the-art on this benchmark but to investigate parsing performance fluctuation when cross-lingual components (gold POS annotations and supervised multilingual embeddings) are replaced by those obtained in an unsupervised manner.

We investigate the following variations to each component of the input representation.

\begin{itemize} \itemsep -1pt
    \item {\bf Multilingual word and cluster embeddings.} The original work of \newcite{malopa} uses `robustly projected' pre-trained embeddings \cite{guo2015cross} for word embeddings and embeddings learned from English Brown cluster IDs projected through word alignments \cite{guo2016representation} for cluster embeddings; both of these rely on parallel data and we refer to them collectively as {\sc guo}. We replace these with monolingual \textit{fastText} embeddings \cite{Q17-1010} projected to a common space using {\sc muse}, the unsupervised method of \newcite{conneau2017word}.   For cluster embeddings we start with {\it fastText} monolingual embeddings trained over Brown cluster ID sequences instead of word tokens ($|C|=256$, the same as in  \citet{guo2016representation}). Then, 
    unsupervised multilingual embeddings are derived, again using {\sc muse}.\footnote{Both cluster and word {\sc muse} embeddings are projected to the corresponding English space.} Note that this approach, which we refer to collectively as {\sc muse}, requires no parallel data. We compare both {\sc muse} and {\sc guo} approaches in Section~\ref{sec:extrinsic} and Table~\ref{table:dep-muse}.
    
    \item {\bf POS tag scheme.} The original work uses gold POS tag data at both train and test time. While realistic to have gold POS info from PLs for training, it is unrealistic to have this data available for new CLs at test time. We thus compare the original {\sc gold} scenario with the realistic {\sc cipher} scenario, where the training data is still gold, but the test POS tags use the method presented in this work. Another realistic scenario dispenses with POS disambiguation except for the trivial distinction of punctuation; for compatibility purposes this is done in both train and test data and is labeled {\sc none}.
\end{itemize}

We investigate all combinations of \{{\sc guo}, {\sc muse}\}-\{{\sc gold}, {\sc cipher}, {\sc none}\}.

\section{Results and Discussion}

\subsection{Labeling and Cipher Grounding}
\label{section:res-cipher}

The results in Table \ref{table:m1-res} are somewhat at odds with those presented in \newcite{stratos2016unsupervised}, but these are done at different operating points; we use different data, the UD-17 tag set instead of the Universal Treebank 12 tag set, and, perhaps most importantly, generate more clusters. 
We further note that to some degree, choosing Brown clusters based on the results in Table \ref{table:m1-res} compromises claims of our approach being fully `unsupervised' for those six languages, however our subsequent experiments on additional languages are truly unsupervised.\\

Table \ref{table:rl-il} presents the intrinsic performance of the cipher grounder over all PL-CL pairs considered.
The difference between the best and the worst performing PL for each CL ranges from 24.62 percentage points for Swahili to 48.34 points for French, and an average difference of 34.5 points among all languages.
The case when PL$=$CL is also presented in Table \ref{table:rl-il} as a reference and provides a reliable upper-bound under zero-resource conditions.
It is worth noting the difference in accuracy when comparing the best performing PL for each CL with its corresponding PL$=$CL upper-bound.
Among all CLs, the best cipher grounder for French (es-fr) gets the closest to its upper-bound with just 4.81 percentage points of difference, followed by the English grounder (fr-en) with 13.53 points of difference. On the other hand, the best Swahili grounder (ar-sw) is the most distant from its upper-bound with 30.45 points of difference.\\

Given such wide performance gaps in the CL set, the choice of a suitable PL becomes crucial for performance; therein the cipher model confidence and typological similarity are explored as possible choice criteria.
With regards to model confidence, the Pearson correlation between accuracy scores and PPL, expected to be negative, ranges from $-0.71$ for English to 0.40 for Farsi.
Since the PPL values for different PLs are not comparable, we first z-normalize PPL per CL and then concatenate the results for all CLs. The Pearson correlation of the resulting PPL-accuracy values is -0.13.
This last result indicates that the most confident model might not be the most accurate, hence this criterion is not suitable for choosing a suitable PL.\\

With regards to typological similarity, we find that the Pearson correlation between accuracy scores and cosine similarity of typological feature vectors, expected to be positive, ranges from 0.44 for English to -0.14 for Farsi. The total correlation is found to be 0.18.
Again, we find that the most typologically similar $s$ might not be the the most accurate, hence this criterion is not suitable either.

Hence, it becomes obvious that choosing a single PL is an inefficient strategy that does not leverage the contribution that other PLs could bring.
In this situation, the combination of cipher grounders for several PLs represents a sound strategy when no prior linguistic information of a certain CL is available. As shown in Table~\ref{table:rl-il}, this model, {\sc cipher-avg}, obtains accuracy scores of 56.4, 58.6, 37.4, and 37.8 \% for en, fr, fa, and sw, respectively.
When compared to the best performing PL for each CL (see bold cells in Table \ref{table:rl-il}), it can be noticed that the performance gap ranges from just 1.2 percentage points for Swahili to 13.3 points for French, with an average of 6.1 points among all target languages.

Let us now compare the performance of {\sc cipher-avg} with that of a vanilla supervised neural model.\footnote{We use UDPipe v1.2.0 \cite{udpipe:2017} to train the models.} Table \ref{table:noun-cipher-sup} shows precision, recall, and F1 scores for the NOUN tag.
Even though {\sc cipher-avg} achieved mixed results (mid to low accuracy), the model robustly achieves mid-range performance according to F1-score for all CLs. 
The results are even more optimistic in terms of recall for English and French, and in terms of precision for Farsi and Swahili.
This gives us hope that  {\sc cipher-avg} can provide a useful, if noisy, signal to downstream tasks that depend on non-trivial performance over specific POS tags, such as name tagging, as exposed in the next section.

\begin{table*}
\begin{tabular}{|c|cccccccccc|c|c|}
\hline
   & \multicolumn{10}{c|}{Parent Language (PL)} &&    \\
CL & en    & de    & fr             & it    & es             & ja    & cs    & ru             & ar             & sw    &{\footnotesize \sc cipher-avg}& PL=CL \\ \hline
en & -     & 57.1 & \textbf{60.4} & 59.9 & 59.4           & 25.1 & 52.8  & 49.0          & 30.7          & 28.4 &56.4& 73.9 \\
fr & 58.1 & 56.0 & -              & 68.6 & \textbf{71.9} & 23.6 & 48.3 & 47.8          & 35.0          & 26.7 &58.6& 76.7 \\
fa & 13.8 & 32.3 & 29.7          & 22.7 & 33.3          & 19.7 & 33.3 & \textbf{43.5} & 37.0          & 38.2 &37.4& 73.3 \\
sw & 24.9 & 14.3 & 37.3          & 21.2 & 35.9          & 21.3 & 25.8 & 27.9          & \textbf{38.96} & -     &37.8& 69.4 \\
\hline

\end{tabular}
\caption{Performance of cipher grounder using {\sc brown } ($|C|=500$) as labeler. The best PL for each CL besides itself, is shown in bold. The artificial case where we have CL POS data (PL$=$CL) is shown for comparison, as is the ultimately used {\sc cipher-avg} method.
}
\label{table:rl-il}
\end{table*}

\begin{table*}
\centering
\begin{tabular}{|c|ccc|ccc|}
\hline
   & \multicolumn{3}{c|}{ {\sc cipher-avg} } & \multicolumn{3}{c|}{Supervised} \\
CL & P       & R       & F1      & P        & R        & F1       \\ \hline
en & 47.70   & 64.4    & 54.81   & 94.04    & 90.44    & 92.20    \\
fr & 56.26   & 78.82   & 65.65   & 96.15    & 93.72    & 94.92    \\
fa & 64.94   & 51.23   & 57.27   & 96.48    & 97.77    & 97.12    \\
sw & 53.46   & 51.82   & 52.63   & 98.88    & 97.50    & 98.18    \\ \hline
\end{tabular}
\caption{Comparison of performance over the NOUN tag, as measured by precision (P), recall (R), and F1 scores, between our combined cipher grounder ({\sc cipher-avg}) and a supervised tagger. }
\label{table:noun-cipher-sup}
\end{table*}

\subsection{Extrinsic evaluation}
\label{sec:extrinsic}

In the name tagging task, our LSTM-CNN baseline obtains $78.76\%$ and $70.76\%$ F1 score for Kinyarwanda and Sinhalese, respectively. When enriching the input representation with {\sc cipher-avg} tags, the performance goes up to $80.16\%$ and $71.71\%$ respectively.
These results suggest that the signal provided by the combined cipher grounder is significant enough for relevant tags such as common, proper nouns and noun modifiers.
As an example, consider the sentence \texttt{Kwizera Peace Ndaruhutse , wari wambaye nomero 11.}
The baseline model fails to recognize \texttt{Kwizera Peace Ndaruhutse} as a person name. In contrast, with the PROPN tag assigned by {\sc cipher-avg} to \texttt{Kwizera}, \texttt{Peace}, and \texttt{Ndaruhutse}, our model is able to identify this name.

Likewise, the utility of {\sc cipher-avg} tags for dependency parsing under zero-resource scenarios is summarized in Table \ref{table:dep-guo} and Table \ref{table:dep-muse}.
It is important to point out that, even though the \malopa~setup follows the no-treebank setup of \citet{malopa}, parsing scores in the first row of Table \ref{table:dep-guo} differ from those reported by them (Table 8 in \citet{malopa}). 
Such difference is to be expected since the underlying parser used in our experiments is a graph-based neural parser \cite{dozat2016biaffine} instead of a transition-based one \cite{dyer2015transition}.\footnote{Due to time constraints, we could not experiment with longer training regimes possibly needed given the high block dropout rates in \citet{dozat2016biaffine}.}
As mentioned earlier, our objective is to analyze the effect of our tagger's signal on parsing performance under no-resource scenarios, instead of pushing the state-of-the-art for the task.

We first analyze the effect of POS tag information at test time for the \malopa~setup in Table~\ref{table:dep-guo}. First we remove all POS signal except trivial punctuation information ({\sc none} row), and, predictably, the scores drop significantly across all target languages.
Then, we use our cipher tags ({\sc cipher} row) and see improvements for all languages in LAS and for all but one language in UAS (de).
This demonstrates the value of our cipher approach.

We then take the next logical step and remove the parallel data-grounded embeddings, replacing them with fully unsupervised {\sc muse} embeddings. Table~\ref{table:dep-muse} summarizes these results.
Let us compare {\sc muse-none} setup (no POS signal at train or test time) with {\sc muse-gold} (gold POS signal at train and test time).
It can be observed that POS signal improves performance greatly for all languages when using {\sc muse} embeddings.
However, consider {\sc guo-gold} and {\sc muse-none}.
Here we note a mixed result: whilst de, sv, and it do benefit from POS information, the other languages do not, obtaining great improvements from {\sc muse} embeddings instead.
Finally, consider {\sc muse-cipher} (gold POS tags during training, cipher tags during testing). When compared to {\sc muse-none} setup, it can be observed that, unfortunately, the heuristic POS tagger is too noisy and gets in {\sc muse}'s way.

\begin{table*}[]
\centering
\scriptsize
\begin{tabular}{|c||rr|rr|rr|rr|rr|rr|} \hline
                        & \multicolumn{2}{c}{de}& \multicolumn{2}{c}{fr} & \multicolumn{2}{c}{es} & \multicolumn{2}{c}{it} & \multicolumn{2}{c}{pt} & \multicolumn{2}{c|}{sv} \\  
 Test Tags & UAS   &  LAS         & UAS     & LAS          & UAS   & LAS            & UAS   & LAS            & UAS   & LAS            & UAS   & LAS \\ \hline
 {\sc gold}              & 65.57  & 52.37         & 71.27  & 59.80        & 73.26 & 63.13          & 71.46 & 59.66          & 63.28 &  54.93         & 77.50 & 64.90 \\ \hline
 {\sc none}              & 40.90  & 18.61         & 51.14  & 30.91        & 43.82 & 17.67          & 48.22 & 33.29          & 37.89 & 16.72         & 38.15 & 17.96 \\
 {\sc cipher} (this work)& 38.31  & \textbf{24.72}         & \textbf{54.46}  & \textbf{41.04}        & \textbf{55.56} & \textbf{41.16}          & \textbf{54.05} & \textbf{39.78}          & \textbf{46.97} & \textbf{36.07}          & \textbf{55.06} & \textbf{36.51} \\ \hline
\end{tabular}
\caption{Impact of grounded unsupervised POS tagging on \malopa's `zero-resource' condition. Bold entries indicate an improvement over the baseline condition of having no POS tag information (beyond punctuation)}
\label{table:dep-guo}
\end{table*}

\begin{table*}[]
\centering
\scriptsize
\begin{tabular}{|cc||rr|rr|rr|rr|rr|rr|}  \hline
     &     & \multicolumn{2}{c}{de}& \multicolumn{2}{c}{fr} & \multicolumn{2}{c}{es} & \multicolumn{2}{c}{it} & \multicolumn{2}{c}{pt} & \multicolumn{2}{c|}{sv} \\
Embeddings & Test Tags & UAS   &  LAS          & UAS     & LAS          & UAS   & LAS            & UAS   & LAS            & UAS   & LAS            & UAS   & LAS \\ \hline  
{\sc guo}  & {\sc gold}    & 65.57 & 52.37    & 71.27 & 59.80 & 73.26 & 63.13  & 71.46  & 59.66& 63.28  & 54.93   & \textbf{77.50}                                                                & \textbf{64.90}                   \\
{\sc muse} &  {\sc gold}  & \textbf{66.19}  & \textbf{56.28}    & \textbf{80.86} & \textbf{72.65} & \textbf{81.06} & \textbf{73.62}  & \textbf{82.08}  & \textbf{72.40}& \textbf{81.17}  & \textbf{76.17}   & 72.46                                                                & 61.71                  \\
{\sc muse} & {\sc none}    & 57.26  & 45.10    & \textit{73.84} & \textit{63.09} & \textit{77.01} & \textit{67.06}  & 71.36  & \textit{60.48}& \textit{75.31}  & \textit{68.36}   & 60.82                                                                & 45.25                  \\
{\sc muse} & {\sc cipher} & 48.56  & 37.13    & 69.94 & 59.22 & 73.86 & 61.68  & 69.30  & 56.85& 73.41  & 65.23   & 57.39                                                                & 41.49                  \\ \hline
\end{tabular}
\caption{Changing to unsupervised {\sc muse} embeddings boosts \malopa's zero-resource performance significantly (\textbf{bold} entries), in many cases doing so even without any POS tag information (\textit{italic} entries), however noisy decipherment-based POS tags are no longer helpful.}
\label{table:dep-muse}
\end{table*}

\section{Related Work}

Our proposed tagging pipeline can be interpreted as first reducing the vocabulary size to a fixed number of
clusters, and then finding a cluster--POS tag mapping table that best explains the data without any path constraint (a cluster ID could be mapped to any POS tag). In this sense, our approach applies EM to simplify the task (e.g.\ when using Brown clustering \cite{brown1992class}), followed by another EM run to optimize cipher table parameters.

Under this lens, the methods closest to our approach are those which attempt to reduce or constrain the parameter search space prior to running EM.
For instance, \citet{ravi2009minimized} explicitly search for the smallest model that explains the data using Integer Programming, and then use EM to set parameter values. In a different approach, \citet{goldberg2008can} obtain competitive performance with a classic HMM model by initializing the emission probability distribution with a mixture of language-specific, linguistically constrained distributions. However, both of these approaches are framed around the task of unsupervised POS {\it disambiguation} with a full dictionary \cite{merialdo1994tagging}.
Previous work relaxes the full dictionary constraint by leveraging monolingual lexicons \cite{haghighi2006prototype,smith2005contrastive,merialdo1994tagging,ravi2009minimized}, multilingual tagged dictionaries \cite{li2012wiki,fang2017model}, and parallel corpora \cite{duong-EtAl:2014:EMNLP2014,tackstrom2013token,das2011unsupervised}.

In addition, previous work includes sequence models that do not rely on any resource besides raw text during training, namely unsupervised POS {\it induction} models. These models are based, with few exceptions, on extensions to the standard HMM; most, in the form of appropriate priors over the HMM multinomial parameters \cite{goldwater2007fully,johnson2007doesn,graca2009posterior}; others, by using logistic distributions instead of multinomial ones \cite{berg2010painless,stratos2016unsupervised}. However, these models still need to ground or map hidden states to actual POS tags to evaluate, and they inevitably resort to many-to-one or one-to-one accuracy scoring. Some previous work has been cautious in pointing out this ill-defined setting \cite{ravi2009minimized,christodoulopoulos2010two}, and we argue its inappropriateness for scenarios in which the test set is extremely small or even when no annotated reference corpus exists.

Therefore, the problem of grounding the sequence of states or cluster IDs to POS tags without using any linguistic resource remains unsolved. We formulate this task as a decipherment problem.
Decipherment aims to find a substitution table between alphabets or tokens of an encrypted code and a known language without the need of parallel corpora. 
The task has been successfully applied in alphabet mapping for lost languages \cite{snyder2010statistical}, and machine translation at the character  \cite{pourdamghani2017deciphering} and token level \cite{ravi2011deciphering,dou2015unifying}. For the task of POS tag grounding, the sequence of states or cluster IDs is modeled as an encrypted code to be deciphered back to a POS sequence.
Furthermore, we tackle the problem from a `universal' perspective by allowing the cipher learn from POS sequences from a varied pool of languages.

Other recent work has declared a `radically universal' mantra to language inclusivity. \newcite{hermjakob-may-knight:2018:Demos} presents a Romanizer that covers all writing systems known to Unicode. \newcite{pan-EtAl:2017:Long2} extends name tagging and linking capability to hundreds of languages by leveraging Wikipedia. \newcite{KIROV16.1077} has semi-automatically built inflectional paradigms for hundreds of languages.

\section{Conclusion}

We present a POS tag grounding strategy based on decipherment that does not require human-labeled data to map states or clusters to actual POS tags and thus can be used in real-world situations requiring grounded POS tags.
The decipherment model considers state or word cluster IDs of a CL as a cipher text to be deciphered back to a POS sequence.

The model operates on top of Brown cluster IDs and requires a POS language model trained on annotated corpora of one or more PLs.
Experimental results over a large and linguistically varied set of PLs show that the choice of which PL to decipher POS tags from is crucial for performance. We explore model confidence, as measured by perplexity and typological similarities, as intuitive criteria for PL choice. However, both criteria prove to be not correlated with tagging accuracy scores.
Thus, we propose a cipher model combination strategy in order to leverage the word-order patterns in several PLs, at the cost of an accuracy drop ranging from just 1.15 percentage points to 13.33 points.

The resulting combined grounder is completely language agnostic, making it attractive for the analysis of languages new to the academic community. Furthermore, analysis over the tasks of name tagging and dependency parsing demonstrate that the tags induced by the combined grounder provide a non-trivial signal for improvement of the downstream task.
We obtain state-of-the-art results for name tagging in Kinyarwanda and Sinhalese, languages for which POS annotated corpora is nearly non-existent.

\section*{Acknowledgments}

Thanks to Xusen Yin, Nima Pourdamghani, Thamme Gowda, and Nanyun Peng for fruitful discussions. This work was sponsored by DARPA LORELEI (HR0011-15-C-0115).

\bibliography{naaclhlt2019}
\bibliographystyle{acl_natbib}

\end{document}